%% file: main.tex
\documentclass[conference]{IEEEtran}
\IEEEoverridecommandlockouts
\usepackage{cite}
\usepackage{amsmath,amsfonts}
\usepackage{algorithmic}
\usepackage{graphicx}
\usepackage{textcomp}
\usepackage{xcolor}
\usepackage[ruled, vlined,linesnumbered]{algorithm2e}
\usepackage{epstopdf}
\usepackage{comment}
\usepackage{multirow}
\usepackage{url}
\usepackage{optidef}

\usepackage{amsthm}
\theoremstyle{definition}

\newcommand{\our}{\textit{FedFM}\xspace}

\makeatletter
\newcommand{\linebreakand}{%
  \end{@IEEEauthorhalign}
  \hfill\mbox{}\par
  \mbox{}\hfill\begin{@IEEEauthorhalign}
}
\makeatother

\def\BibTeX{{\rm B\kern-.05em{\sc i\kern-.025em b}\kern-.08em
    T\kern-.1667em\lower.7ex\hbox{E}\kern-.125emX}}
\begin{document}

\title{\our: Towards a Robust Federated Learning Approach For Fault Mitigation at the Edge Nodes}

\author{\IEEEauthorblockN{Manupriya Gupta}
\IEEEauthorblockA{\textit{CSE Department, IIT Delhi}\\
New Delhi, India \\
guptamanupriya.iitd@gmail.com}
\and
\IEEEauthorblockN{Pavas Goyal}
\IEEEauthorblockA{\textit{CSE Department, IIT Delhi}\\
New Delhi, India \\
pavasgdb@gmail.com}
\and
\IEEEauthorblockN{Rohit Verma}
\IEEEauthorblockA{\textit{University of Cambridge}\\
Cambridge, UK \\
rv355@cam.ac.uk}
\linebreakand
\IEEEauthorblockN{Rajeev Shorey}
\IEEEauthorblockA{\textit{UQIDAR, IIT Delhi}\\
New Delhi, India \\
rshorey@iitd.ac.in}
\and
\IEEEauthorblockN{Huzur Saran}
\IEEEauthorblockA{\textit{CSE Department, IIT Delhi}\\
New Delhi, India \\
saran@cse.iitd.ac.in}
}

\maketitle

\begin{abstract}
Federated Learning deviates from the norm of \textit{"send data to model"} to \textit{"send model to data"}. When used in an edge ecosystem, numerous heterogeneous edge devices collecting data through different means and connected through different network channels get involved in the training process. Failure of edge devices in such an ecosystem due to device fault or network issues is highly likely. In this paper, we first analyse the impact of the number of edge devices on an FL model and provide a strategy to select an optimal number of devices that would contribute to the model. We observe how the edge ecosystem behaves when the selected devices fail and provide a mitigation strategy to ensure a robust Federated Learning technique.
\end{abstract}


\input{Introduction}
\input{RelatedWork}
\input{WorkerAnalysis}
\input{FaultAnalysis}
\input{FaultMitigation}
\input{Evaluation}
\input{Discussion}
\input{Conclusion}

\bibliographystyle{IEEEtran}
\bibliography{ref}

\end{document}

%% file: Introduction.tex
\section{Introduction}~\label{introduction}
Traditional, centralised machine learning (ML) involves training on large datasets available on a central server. Once trained, these models are either deployed on edge devices or receive data to infer the result. Centralised ML essentially requires data collected from all devices to be made available to the training process at the server. However, with the growing privacy concern about sharing data with a centralised server and the development of edge devices with more computationally efficient processors, Federated Learning (FL) was introduced by Google~\cite{45648} as an alternative. In contrast to the traditional ML, Federated Learning sends the model to the edge devices where the data is available, instead of asking for the data from these edge devices (known as \textit{workers or worker nodes}). All available edge devices perform local training and send the trained \textit{Local model} to a centralised aggregator that generates a \textit{Global model} based on the \textit{Local models} available from all edge devices. This \textit{Global model} is then sent to all the edge devices. This process is repeated multiple times to improve the model further until a \textit{convergence} is obtained. Two crucial parameters define the efficiency of an FL technique; these are \textit{Convergence Time ($\mathcal{C}$)}, which is the time it takes for the learning model to converge and the \textit{Model Accuracy ($\mathcal{A}$)}~\cite{nguyen2021federated}.

As is evident from the workflow of FL, the availability of edge devices plays a crucial role. However, there is a high probability of workers failing in this ecosystem of heterogeneous edge devices connected through different network channels. The failure could either be because of a fault with the edge device or network disruption~\cite{bonawitz2019towards}. Such failures, especially during a training iteration, are likely to adversely affect $\mathcal{C}$ and $\mathcal{A}$. A majority of the state-of-the-art FL models~\cite{wang2019adaptive,yu2018parallel} discuss convergence time guarantees, but they do not discuss how the model would perform in such failure scenarios. A few works do analyse less number of workers in the ecosystem~\cite{smith2017federated,li2018federated}. However, these works again do not consider the impact of workers failing during the training phase. An acceptable state-of-the-art approach is to ignore such device failures and work with the remaining workers~\cite{bonawitz2019towards}. However, if a failing worker has an exclusive data sample, then the trained model will be biased towards the other data. Another strategy employed by some models is to introduce replications in the ecosystem~\cite{tao2020straggler}. Unfortunately, replication would involve sharing data between one or more workers, defeating the purpose of using FL to ensure privacy.

\begin{figure*}[!ht]
  \centering
  \includegraphics[width=1\linewidth]{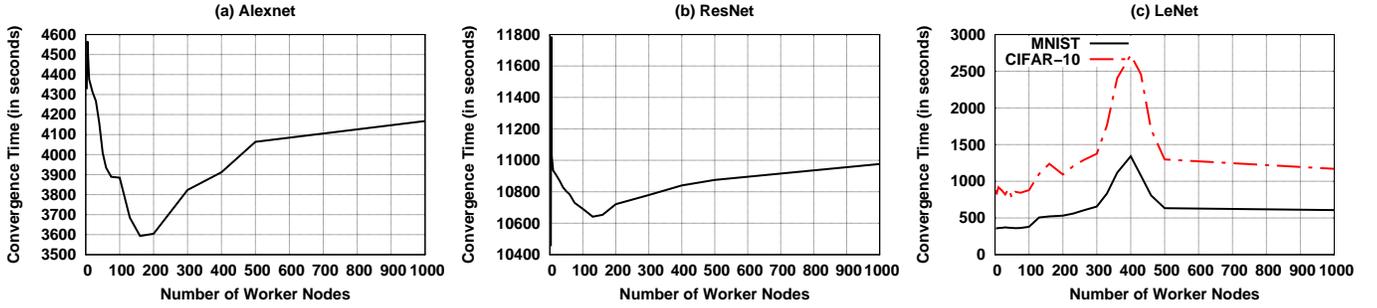}
  \caption{Impact of Worker Count on the Convergence Time for Different Learning Models.}
  \label{fig:worker_count}
\end{figure*}

Designing a robust FL technique thus asks for mitigation steps when such faults occur in the edge ecosystem. However, such robust technique would require analysis of how the workers in the ecosystem impact the learning model parameters. Furthermore, different learning models have different architectures, which could also impact workers' influence on the FL technique. Moreover, not all workers would be required to reach the \textit{convergence}. Understanding the impact of the failure of the worker nodes is crucial for the development of a robust federated learning technique. In this paper, we try to address the above points. The key contributions are as follows:
\begin{itemize}
    \item Analyse the impact of the number of workers on different deep neural network models and different datasets in a federated learning environment.
    \item Design a strategy to select the optimal number of worker nodes to ensure low $\mathcal{C}$ and high $\mathcal{A}$.
    \item Analyse the impact of failure of workers on $\mathcal{C}$ and $\mathcal{A}$.
    \item Design a federated learning technique, \textit{Federated Fault Mitigation} (\our), that incorporates fault mitigation in the training process.
\end{itemize}

In the next section, we provide a background of Federated Learning and the state-of-the-art works ($\S\ref{relatedwork}$). The following sections are dedicated to analysing the impact of worker nodes, designing the optimal worker selection strategy ($\S\ref{workeranalysis}$), analysing the impact of worker node failures ($\S\ref{faultanalysis}$), and incorporating fault mitigation strategy in model training ($\S\ref{faultmitigation}$). We evaluate \our in a simulation and a prototype environment highlighting its efficiency ($\S\ref{evaluation}$). Following this, we provide a discussion of future directions ($\S\ref{discussion}$) and then the final concluding remarks ($\S\ref{conclusion}$).

%% file: RelatedWork.tex
\section{Related Work}~\label{relatedwork}
Edge computing is defined as a part of a distributed computing topology in which information processing is located close to the edge, where things and people produce or consume that information. With recent advancements in edge computing, the need has arisen to bring AI to the edge~\cite{satyanarayanan2017emergence}. However, in contrast to the centralized AI techniques, which require edge devices sharing data to the server to ensure privacy, AI on edge would instead move towards sharing model to the edge devices. Privacy concerns paved the way for the introduction of Federated Learning~\cite{45648} that enhances security in edge ecosystems by not mandating the edge devices to share any data to a central server. 

Federated Learning is a decentralized learning mode, making learning highly personalized for users involving very low latency and preserving privacy. A standard Federated Learning network architecture consists of the edge devices (worker nodes) and the central server (aggregator). The worker nodes perform local training on their local dataset and send the local model to the aggregator. On receiving local models from the worker nodes, the aggregator utilizes an aggregation algorithm to generate a global model. Several such aggregation algorithms have been developed to serve different purposes, The earliest ones being FedSGD~\cite{shokri2015privacy}, which is a direct transposition of the Stochastic Gradient Descent algorithm and FedAvg~\cite{mcmahan2017communication} that instead of using gradients uses the average of local model weights to generate the global model. The aggregation of the local models are done over multiple iterations that are concluded once the model reaches a convergence. Convergence in most techniques is defined as a point when the training loss crosses a certain threshold or becomes stable with no visible change for a while.

An overall efficient Federated Learning model also needs to be efficient in scenarios with a heterogeneous distribution of resources, and some nodes might fail. Increasing the number of nodes will help us in increasing accuracy at the cost of latency. What type of mitigation strategy would be best in this case? Wang et al.~\cite{wang2019edge} show the potential of integrating Deep Learning and Federated Learning in an Edge computing environment and how edge computing is more efficient in handling the heterogeneous node capabilities than other networks. Several works talk about node capabilities when we have the scenarios of heterogeneous resource allocation~\cite{jeon2020optimal,nishio2013service,sardellitti2015joint}. However, these are only aimed at minimizing the computational time and power and might compromise on latency. Similarly, Konecny et al.~\cite{konevcny2016federated} present compression algorithms to achieve lower communication latency at the cost of sacrificing accuracy. Wang et al.~\cite{wang2018edge} also try to tune the parameters involved in FL, such as the number of epochs in given resource constraints, but do not take into account the heterogeneous computation or data resources of nodes which is true in real-life scenarios.

A majority of the Federated Learning techniques only consider fault tolerance but do not give steps to mitigate in case faults occur~\cite{wang2019adaptive,yu2018parallel,javed2020iotef}. Javed et al.~\cite{javed2020iotef}, for instance, employ a Kubernetes-based fault tolerance scheme for federated learning but does not discuss mitigation. Another class of works discuss the impact of low worker node population, but even these works never talk about the impact on the FL technique when nodes start failing~\cite{smith2017federated,li2018federated}. Other works ignore the worker node when it fails and continue the process with the other working nodes~\cite{bonawitz2019towards}. Ignoring nodes introduces bias in the learning model and hence should not be considered a suitable approach. Some works perform fault mitigation, but they rely on data replication~\cite{tao2020straggler}. However, data replication essentially defeats the purpose of data privacy, for which FL came into the picture.

%% file: WorkerAnalysis.tex
\section{Worker Analysis}\label{workeranalysis}
In this section, we analyse the impact of the number of worker nodes on the FL technique and, assuming a non-failure case, compute what percentage of worker nodes would be optimal to ensure low $\mathcal{C}$ and high $\mathcal{A}$.

\subsection{Worker Count Impact on FL Model}\label{subsec:workercount}
In the first set of experiments, we study the impact of the number of worker nodes (hereafter the set of all worker nodes is represented as $\mathcal{K}$) on \textit{Convergence Time}. We define \textit{Convergence Time} as the total time taken by the edge ecosystem to collectively reach a constant training loss while ensuring that the training loss is below a certain threshold. The threshold is decided such that the \textit{Model Accuracy}, which is the test accuracy, never falls below the state-of-the-art test accuracy of the learning model for a given dataset. However, there could be training instances, when the model could not meet the state-of-the-art accuracy threshold. In such a scenario, the training loss becomes constant at a higher value (lower $\mathcal{A}$) and hence, the \textit{Convergence Time} is taken as the time when this higher constant training loss is observed.

We set up a simulation environment with Pysyft~\cite{pysyft}. The simulations are run on an Ubuntu 20.04 system~\cite{ubuntu} that has 12 GB RAM, an octa-core 1.5 GHz processor, and 16 GB Nvidia T4 GPU. We use a centralised FL architecture, where the aggregator is fixed, and all workers send their local models to the aggregator to generate the global model. Pysyft creates virtual workers, and the aggregator has a pointer to the tensors that are on the workers. Each worker and the aggregator are assigned a separate IP address. The simulation environment also provides the capability to specify the time it takes for a network transmission to occur. We utilise this property to simulate the time a worker node with a certain bandwidth would take to transmit a local model to the aggregator.

We run our experiments with three different learning models. These are LeNet~\cite{lecun1998gradient}, AlexNet~\cite{krizhevsky2012imagenet}, and ResNet~\cite{he2015deep} and two popular datasets MNIST~\cite{lecun1998mnist} and CIFAR-10~\cite{krizhevsky2009learning}. We vary the number of worker nodes from 1 to 1000 in our experiments.

% \begin{figure}[!ht]
%   \centering
%   \includegraphics[width=1\linewidth]{images/competing.eps}
%   \caption{Impact of Worker Count on the Convergence Time for AlexNet with MNIST Dataset.}
%   \label{fig:worker_count}
% \end{figure}

% \begin{figure}[!ht]
%   \centering
%   \includegraphics[width=1\linewidth]{images/competing.eps}
%   \caption{Impact of Worker Count on the Convergence Time for ResNet with MNIST Dataset.}
%   \label{fig:worker_count}
% \end{figure}

% \begin{figure}[!ht]
%   \centering
%   \includegraphics[width=1\linewidth]{images/competing.eps}
%   \caption{Impact of Worker Count on the Convergence Time for LeNet with MNIST and CIFAR-10 Dataset.}
%   \label{fig:worker_count}
% \end{figure}

Figure~\ref{fig:worker_count} shows the variation of $\mathcal{C}$ when the number of worker nodes increases from 1 to 1000. We observe two trends when we increase the number of worker nodes. AlexNet and ResNet show a convex behaviour, while LeNet shows a concave behaviour. $\mathcal{C}$ of an FL technique depends on multiple factors. These factors could be the capability of the worker node, type of communication, dataset volume on the worker node, client interaction, and the architecture of the learning model~\cite{hao2020time,dhakal2019coded}. It is to be noted that the above experiments are run with a similar environment for all worker nodes and homogeneous data distribution. Hence, the trend could be linked to how these models are executed in an FL environment. As can be seen in Figure~\ref{fig:worker_count}(c), for both datasets, LeNet takes comparable $\mathcal{C}$ even with 1000 workers. However, for AlexNet and ResNet, when increasing the number of workers, $\mathcal{C}$ increases considerably for the CIFAR-10 dataset~\footnote{We only show MNIST results for AlexNet and ResNet because the convergence time is too high when running for the CIFAR-10 dataset.}. As a comparison, for three nodes, AlexNet took $5.5$ hours to converge with CIFAR-10 compared to $1.2$ hours with MNIST. Hence, for simpler models like LeNet that converge faster, inclusion of more worker nodes incurs redundant aggregation leading to higher values of $\mathcal{C}$ with much lesser number of nodes compared to the other two models. 

The key takeaway from these experiments is that the number of worker nodes is crucial for the FL technique. We next perform experiments to investigate the percentage of worker nodes that would suffice for the efficient working of an FL technique.

\subsection{Optimal Worker Percentage for Federated Learning}
Several existing works have shown that local models from all existing workers in the edge ecosystem need not be required to obtain an efficient model~\cite{cho2020client,jeon2020optimal,nishio2019client}. In the next set of experiments, we show that relying on a lesser number of workers improves $\mathcal{C}$ considerably without compromising on $\mathcal{A}$. Following this, we also devise a strategy to select an optimal set of worker nodes to ensure low $\mathcal{C}$ and high $\mathcal{A}$.

We use a similar simulation environment as described in Section~\ref{subsec:workercount} to compute the impact of the number of nodes selected on $\mathcal{C}$ and $\mathcal{A}$. In this and all the following experiments, we use a simple four-layer CNN (called MCNN henceforth) with two Convolution layers and two Fully Connected layers. This model is similar to the standard state-of-the-art models~\cite{mcnn} used to train over the MNIST dataset, and we observe that MCNN gives a test accuracy of $97\%$ when run on the MNIST test data. Using MCNN for the experiments also ensures that the run time of the experiments is lower as compared to the models in Figure~\ref{fig:worker_count}. We run MCNN with 50, 100 and 200 nodes.

\begin{figure}[!ht]
  \centering
  \includegraphics[width=1\linewidth]{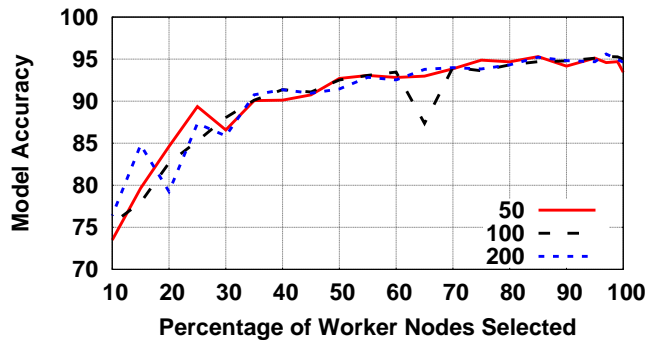}
  \caption{Variation of Model Accuracy with a percentage of worker nodes selected.}
  \label{fig:worker_percentage_acc}
\end{figure}

\begin{figure}[!ht]
  \centering
  \includegraphics[width=1\linewidth]{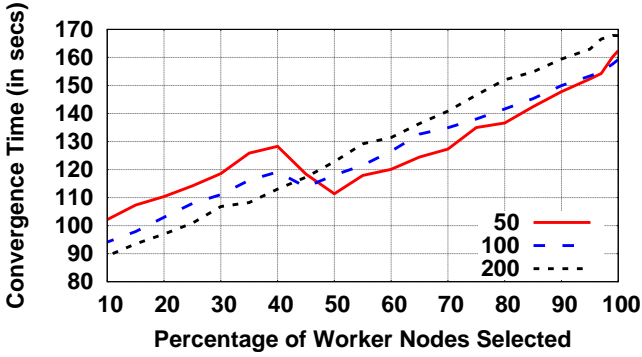}
  \caption{Variation of Convergence Time with a percentage of worker nodes selected.}
  \label{fig:worker_percentage_time}
\end{figure}

We run the FL technique by selecting $10\%$ to $100\%$ of worker nodes for the three node sets and observe the impact on $\mathcal{C}$ and $\mathcal{A}$. All the worker nodes have similar network and system characteristics and follow the same data distribution. Figure~\ref{fig:worker_percentage_acc} and Figure~\ref{fig:worker_percentage_time} show the variation of $\mathcal{A}$ and $\mathcal{C}$ with the varying percentage of worker nodes selected to contribute to the global model, averaged over $100$ runs. We observe in Figure~\ref{fig:worker_percentage_acc} that at $70\%$ of worker nodes, $\mathcal{A}$ is almost similar to what it is at $100\%$. On the contrary, the same $70\%$ of nodes require a $\mathcal{C}$ almost $25\%$ less than what it takes when using all worker nodes (Figure~\ref{fig:worker_percentage_time}). It should be noted that although $\mathcal{C}$ is lower for less number of nodes, the $\mathcal{A}$ also falls as low as $75\%$. This is because, the local model does not converge beyond a certain training loss after multiple iterations and hence has a low $\mathcal{C}$. 

Based on the results, we conclude that for only $70\%$ of worker nodes contributing to the global model, $\mathcal{C}$ is considerably reduced ($\approx 25\%$) with minimal decrease in $\mathcal{A}$. Hereafter, for all our experiments we will be using $70\%$ of the total worker nodes to contribute to the training process. Next we devise a strategy to compute how these $70\%$ of the worker nodes should be selected to obtain the best training performance.

\subsection{Optimal Worker Set Selection Strategy}\label{optimalworkerset}

In the previous experiment, all worker nodes have similar system and network characteristics along with same data distribution. However, in a practical scenario, this is quite unlikely. Different worker nodes would have different \textit{storage capacity}, \textit{computation power}, and \textit{network bandwidth} to connect with the aggregator. Furthermore, the distribution of data would be different at each of the worker nodes. Hence, we need to design a strategy to ensure that the selected $70\%$ of worker nodes would improve $\mathcal{C}$ and $\mathcal{A}$.

Among the three worker node specific parameters, i.e., \textit{storage capacity}, \textit{computation power}, and \textit{network bandwidth}, the storage capacity along with data distribution could be inherently linked to the amount of data available at the worker node. Hence, we use \textit{data volume} ($\mathcal{V}$), \textit{computation power} ($\mathcal{P}$), and \textit{network bandwidth} ($\mathcal{B}$) as the parameters to compute the optimal worker set. 

We first observe how the three parameters work in-silo. We therefore select the top $70\%$ worker nodes that have the highest $\mathcal{V}$ when inferring the impact of data volume on the FL model. Similarly, we select the top $70\%$ worker nodes for $\mathcal{P}$ and $\mathcal{B}$. We also use a random selection strategy to compare with the three strategies.

We perform the experiments with 100 worker nodes and train the MNIST dataset with MCNN. We distribute the data between the worker nodes following a Normal distribution~\cite{feller1968introduction}. This ensures that a fraction of nodes have high data volume, and a similar fraction of nodes have low data volume. The bandwidth is also defined for the worker nodes using Normal distribution.

\begin{figure}[!ht]
  \centering
  \includegraphics[width=1\linewidth]{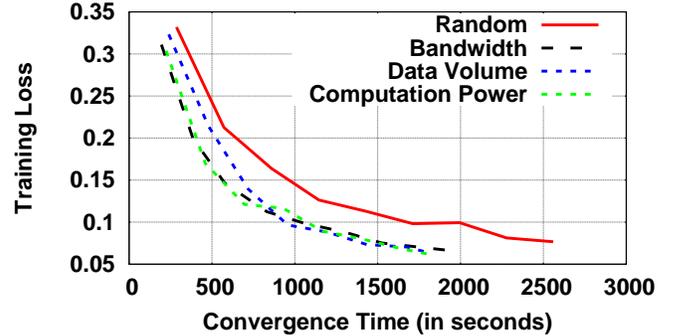}
  \caption{Convergence Time for the FL model when the top $70\%$ nodes are selected using naive strategies.}
  \label{fig:convergence_strategy_naive}
\end{figure}

Figure~\ref{fig:convergence_strategy_naive} shows the time that the selected nodes take to converge to a certain loss value. As is expected, the random selection strategy performs poorly owing to the varying system and network characteristics of the worker nodes. However, not much difference is seen with the other three strategies. The similar trend confirms that all three parameters are crucial when deciding the optimal set of worker nodes for the FL technique. Hence, we need to design a strategy that considers all three parameters.

The impact of $\mathcal{P}$ and $\mathcal{B}$ is quite straightforward as higher computation power would imply a better computation time and a higher bandwidth would result in faster network transmission, thus improving $\mathcal{C}$. However, $\mathcal{V}$ has both positive and negative impact on the model efficiency. A higher $\mathcal{V}$ would mean that the worker node has a higher probability of having useful data samples for training the model. However, a higher value of $\mathcal{V}$ also implies that the worker node needs to perform more computation to generate the local model. Additionally, as seen from Figure~\ref{fig:worker_count}, the learning model plays a crucial role in the efficiency of the FL technique. The learning model parameters like the kernel size, tensor size, number of model parameters, all define how much computation is required and what would be the size of the local model. Hence, the selection strategy should include factors defining the learning model parameters. Observed closely, these learning model parameters are linked to $\mathcal{P}$, $\mathcal{V}$, and $\mathcal{B}$.  The amount of computation required by the learning model is linked with $\mathcal{V}$ and $\mathcal{P}$ of the worker node. While the local model size is linked with $\mathcal{B}$. This is due to the fact that the local model size defines how much time it would take to transmit the local model to the aggregator. 

Effectively, there are three components involved when designing a selection strategy of the optimal worker set. First, the local model transmission time, which is a function of the local model size and $\mathcal{B}$. Second, the computation time on the worker node, which is defined by $\mathcal{V}$, $\mathcal{P}$, and the amount of computation required by the learning model. Finally, the local model training time would depend on $\mathcal{V}$. Higher the $\mathcal{V}$, higher the time to train the local model. Hence, we arrive at the following equation of the \textit{Selection Score} ($\mathcal{S}$), that determines the top $70\%$ worker nodes:
\begin{equation}\label{eqn:selectionstrategy}
    \mathcal{S} = \left( \frac{\alpha}{\mathcal{B}} + \frac{\kappa * \mathcal{V}}{\mathcal{P}} \right) * \frac{1}{\mathcal{V}}
\end{equation}

Here, $\alpha$ and $\kappa$ are constants specific to the learning model. $\alpha$ is computed as the model output size and $\kappa$ is computed as the total memory requirement of the learning model~\cite{params}. Equation~\ref{eqn:selectionstrategy} can be divided into three parts. The first part takes care of the impact of the local trained model size and the network bandwidth ($\frac{\alpha}{\mathcal{B}}$). The second part considers how the computation requirements of the learning model, the available data volume on the worker node and the worker node's computation power impact the selection score ($\frac{\kappa * \mathcal{V}}{\mathcal{P}}$). Finally, the inverse proportionality of the data volume on the selection is captured by the third part ($\frac{1}{\mathcal{V}}$). A higher value of $\mathcal{S}$ would imply a higher possibility of a node to be selected in the optimal set of the top $70\%$ of the worker nodes. 

\begin{figure}[!ht]
  \centering
  \includegraphics[width=1\linewidth]{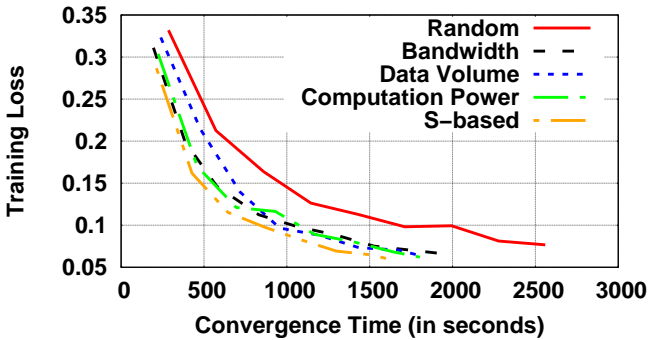}
  \caption{Convergence Time of the FL model when the top $70\%$ nodes are selected for all five selection strategies.}
  \label{fig:convergence_strategy}
\end{figure}

We run a similar experiment as above to calculate the convergence time of the FL technique when the top $70\%$ of the nodes are selected based on Equation~\ref{eqn:selectionstrategy}. It is evident from Figure~\ref{fig:convergence_strategy}, that $\mathcal{S}-based$ selection strategy converges much faster than the other naive strategies. For example, for a training loss of $0.15$, which gives $95\%$ of $\mathcal{A}$, $\mathcal{C}$ is $~1000$ seconds faster than the random strategy and $~300$ seconds faster than the naive strategies. This value of $\mathcal{A}$ is same as what we get in Figure~\ref{fig:worker_percentage_acc} when selecting $70\%$ of nodes. Hence, we observe that even for a heterogeneous data distribution for worker nodes having different system and network characteristics, the $\mathcal{S}-based$ selection strategy ensures that after selecting the top $70\%$ of nodes, $\mathcal{A}$ remains the same as for a homogeneous data distribution over identical nodes.

In the next section, we observe how failure of these selected set of worker nodes affects the efficiency of the FL Model.

%% file: FaultAnalysis.tex
\section{Worker Failure Analysis}\label{faultanalysis}

In the previous section, we came up with a strategy to select top $70\%$ of worker nodes in $\mathcal{K}$ that would be contributing in generating the global model (hereafter referred as $\eta$). In this section, we analyse how the edge ecosystem reacts when a percentage of the worker nodes in $\eta$ start failing.

We design a similar simulation experiment as in Section~\ref{optimalworkerset}. However, we run the experiment with 50, 100, and 200 worker nodes training on MNIST dataset using MCNN. Over $100$ runs of the simulation experiment, we first analyse how average $\mathcal{C}$ and $\mathcal{C}$ vary when a certain percentage of worker nodes fail in the edge ecosystem. We follow a strategy where we select a random subset of worker nodes in $\eta$ and disconnect their connection with the aggregator to simulate failure. Failing only the nodes in the set $\eta$ is important, because the other worker nodes in $\{ \mathcal{K} - \eta \}$ do not contribute to the global model.

\begin{figure}[!ht]
  \centering
  \includegraphics[width=1\linewidth]{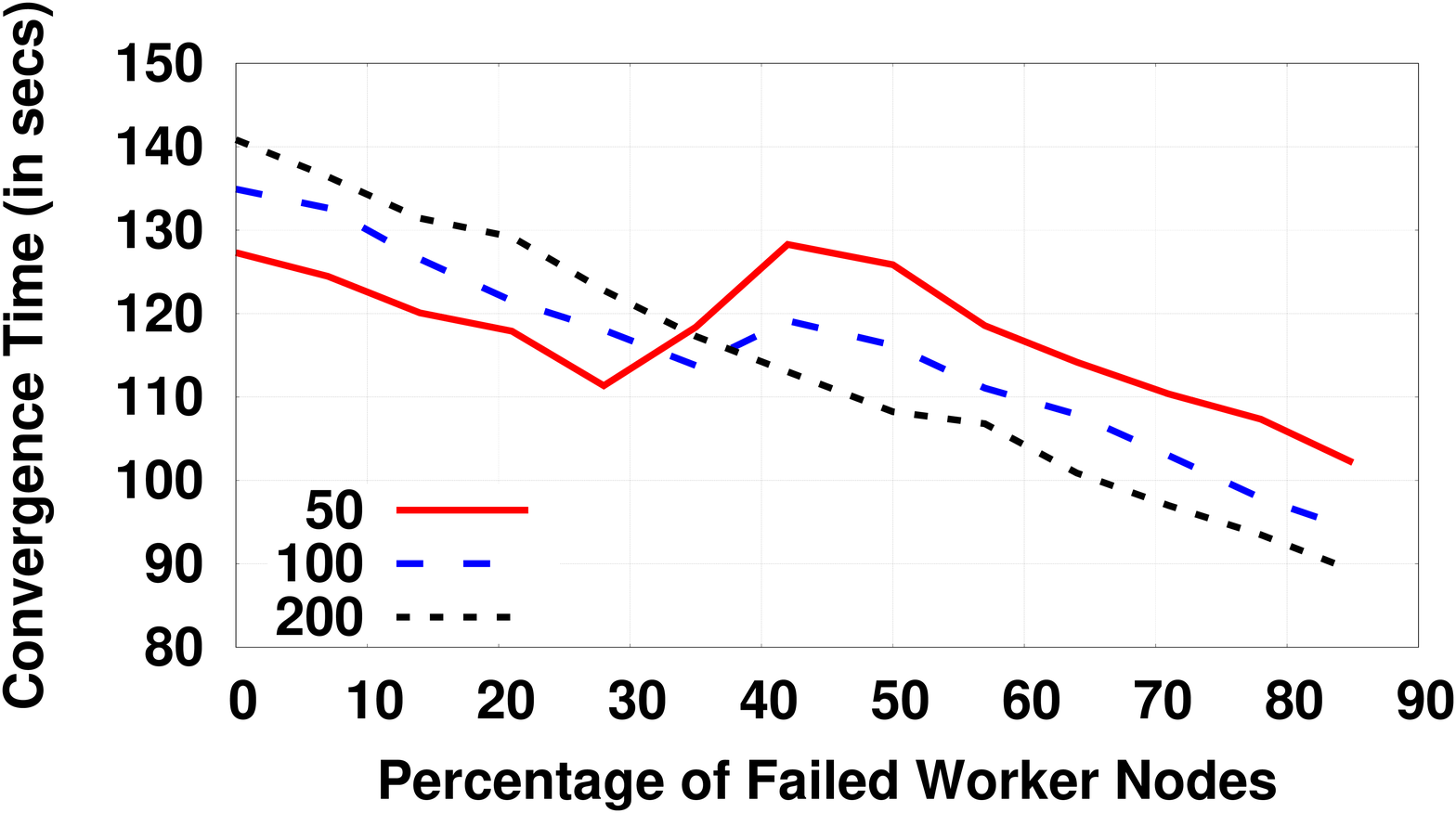}
  \caption{Convergence Time for the FL model a percentage of worker nodes in $\eta$ fail.}
  \label{fig:c_nodefail}
\end{figure}

\begin{figure}[!ht]
  \centering
  \includegraphics[width=1\linewidth]{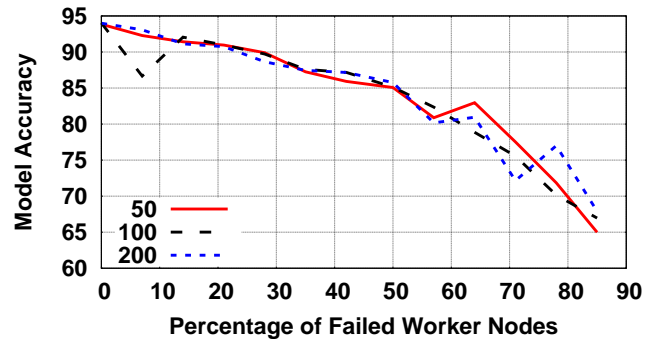}
  \caption{Accuracy of the FL model when a percentage of worker nodes in $\eta$ fail.}
  \label{fig:a_nodefail}
\end{figure}

Figure~\ref{fig:c_nodefail} shows the change in $\mathcal{C}$ when a percentage of the worker nodes in $\eta$ fail. We observe that as the percentage of failed worker nodes increases, $\mathcal{C}$ decreases. This is due to the fact that the learning model does not converge to the state-of-the-art accuracy for the given model. The training loss becomes constant at a higher value resulting in lower convergence time and lower accuracy. This is evident from Figure~\ref{fig:a_nodefail}, where $\mathcal{A}$ decreases when the percentage of failed worker nodes increases.

Consider two scenarios with $\mathcal{K}_1$ and $\mathcal{K}_2$ worker nodes in two edge ecosystem ($\mathcal{K}_1 < \mathcal{K}_2$) having all other equivalent conditions. The set of worker nodes selected after applying the selection strategy are represented as $\eta_1$ and $\eta_2$. Now assume $\nu$ nodes in $\eta_2$ fail, such that, $|\eta_1| = |(\eta_2 - \nu)|$. Although the intuitive understanding would be that considering similar conditions, $\eta_1$ and $(\eta_2 - \nu)$ nodes would perform similarly, it is important to note that this is not true. We perform a set of experiments to investigate this assertion. We run MCNN on MNIST with $|\eta_1|$ varying from 10 to 70. Next, we again run MCNN on MNIST with $|\eta_2| = 70$ nodes. However, for this run, $\nu$ is varied from 60 to 0, such that $|\eta_1| = |(\eta_2 - \nu)|$. For both scenarios we compute the model accuracy for $100$ runs.

% It is important to note that a certain number of contributing worker nodes with no failure is not the same as the same number of contributing worker nodes post failure. For example, an FL model having $50$ contributing worker nodes \textbf{would} not behave the \textbf{same} way as an FL model that had $70$ worker nodes but $20$ of these nodes fail to bring down the contributing worker nodes to $50$. In order to \textbf{prove} this, we perform an experiment where we run MCNN on MNIST with a varying number of worker nodes with no-failure. We run MCNN for 100 nodes, where the top $70\%$ nodes are selected and then compute the \textit{Model Accuracy} for scenarios when a certain number of worker nodes fail, such that we have the same number of working nodes as in the no-failure case.

\begin{figure}[!ht]
  \centering
  \includegraphics[width=1\linewidth]{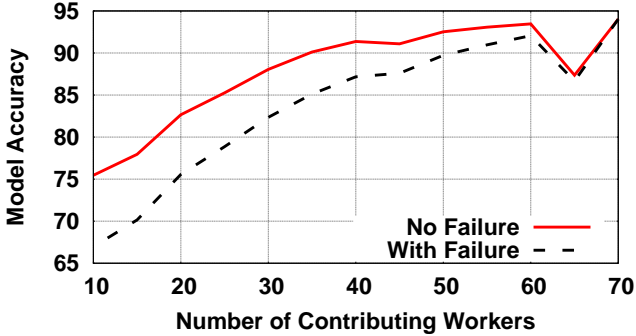}
  \caption{Accuracy of the FL model for the same number of contributing worker nodes for failure and no-failure cases.}
  \label{fig:fail_nofail}
\end{figure}

In Figure~\ref{fig:fail_nofail} we plot $\mathcal{A}$ versus the number of contributing worker nodes for both the scenarios. As observed, the scenario with $\mathcal{K}_1$ worker nodes without any failure has a higher $\mathcal{A}$ than the scenario with failing nodes. The lower accuracy for the scenario where nodes fail is due to the fact that the failed nodes in $\nu$ might have some crucial data samples which when removed due to worker node failure reduces $\mathcal{A}$. On the contrary, all nodes in $\mathcal{K}_1$ successfully contribute to the global model and hence show higher accuracy. This result is the basis of the fault mitigation strategy that we discuss in the next section.

% \textbf{This is intuitive as the selected $70$ nodes had certain data samples which were contributing to the better accuracy and losing these data samples due to node failure would result in a drop in the \textit{Model Accuracy}.} On the contrary, in the non-failure scenario, the edge ecosystem already has the selected set of nodes required to attain a certain accuracy which it always receives. 

%% file: FaultMitigation.tex
\section{\our: A Federated Learning Technique for Fault Mitigation}\label{faultmitigation}

The result that we observed in Figure~\ref{fig:fail_nofail} shows that having all worker nodes contributing always ensures a higher accuracy. This result forms the basis of the mitigation strategy which is triggered on worker node failure at the aggregator.

The aggregator that we use in our previous experiments runs on a central server and follows the Federated Averaging (FedAvg) algorithm~\cite{mcmahan2017communication}. The FedAvg algorithm is a generalisation of the popular Federated Stochastic Gradient Descent (FedSGD) algorithm~\cite{shokri2015privacy} and allows the worker nodes to perform multiple batch updates on its own data. Unlike FedSGD, FedAvg then exchanges the updated weights rather than the gradients. Effectively, the weight obtained from all local models are averaged to obtain global weights for the global model.

\begin{algorithm}
\SetAlgoLined
\KwResult{The Global Federated Learning Model with weight $\omega_{t+1}$}

 $\omega_0 \gets$ initialized model weights
 
 $\mathcal{W} \gets 0.7$ \tcp{Fraction of total nodes to be selected}
 
 $\mathcal{F} \gets \{\}$
 
\ForEach{round $t \in 1,2,...$}{%
    $m \gets max(\mathcal{W}*\mathcal{K}, 1)$\;
    $\mathcal{N}_t \gets$ Select top $m$ workers based on $\mathcal{S}$.\;
    \ForEach{client $k \in \mathcal{N}_t$ in \textbf{parallel}}{%
        $\omega_{t+1}^k \gets$ \textit{ClientUpdate} ($k, \omega_t$)\;
        \If{$\omega_{t+1}^k = null$ after time $\mathcal{T}$}{%
            Append $k$ to $\mathcal{F}$\;
        }
    }
    \If{$|\mathcal{F}| > 0$ and $m > 1$}{%
        $\mathcal{N}_t^f \gets$ Select top $|\mathcal{F}|$ workers based on $\mathcal{S}$.\;
        \ForEach{client $k \in \mathcal{N}_t^f$ in \textbf{parallel}}{%
            $\omega_{t+1}^k \gets$ \textit{ClientUpdate} ($k, \omega_t$)\;
        }
    }
    $\omega_{t+1}^k \gets \sum_{k+1}^m \frac{n_k}{n} \omega_{t+1}^k$
}

 \caption{Federated Fault Mitigation Algorithm (\our) run on the Aggregator. \textit{ClientUpdate ($k, \omega$)}~\cite{mcmahan2017communication} is the same function used by FedAvg.}
 \label{algo:faultmitigation}
\end{algorithm}

We modify the FedAvg algorithm to support the fault mitigation strategy (Algorithm~\ref{algo:faultmitigation}). Given that the edge ecosystem has $\mathcal{K}$ number of worker nodes, the fraction of workers used for contributing to the aggregation process ($\mathcal{W}$) is set as $0.7$. During initialisation, the initial model weights are set as $w_0$ which is sent to all the $\mathcal{K}$ worker nodes. During each iteration, the aggregator decides the number of worker nodes ($m$) which should be selected to contribute in aggregation. Then following the $\mathcal{S}-based$ selection strategy, $m$ nodes are shortlisted. On all of these workers, the \textit{ClientUpdate} function~\footnote{The \textit{ClientUpdate} function is the same as that defined in FedAvg~\cite{mcmahan2017communication} and hence not included in this paper.} is run in parallel. The workers update their own local model weights using \textit{ClientUpdate} and send the updated weights to the aggregator. The aggregator waits for $\mathcal{T}$ time units to receive a response from a client. If no response is obtained after $\mathcal{T}$ time units, the client is added to the failure worker node list $\mathcal{F}$. It should be noted that the threshold $\mathcal{T}$ is configurable and could be changed based on the application use case. After $\mathcal{T}$ time units, if $\mathcal{F}$ is not empty, then the next top $|\mathcal{F}|$ worker nodes are selected from the remaining $|\mathcal{K}| - m$ nodes to form the set $N_t^f$. \textit{ClientUpdate} is again requested on these new nodes. If $|\mathcal{F}|$ becomes greater than $|\mathcal{K}| - m$, then all remaining nodes are included in the new set $N_t^f$. Once updated weights are available from these workers, the global weight is updated following the FedAvg averaging formula (in Line 19 of Algorithm~\ref{algo:faultmitigation}). $n_k$ is the size of local data samples at worker node $k$ and $n$ is the total data sample size. The new averaged weight is then made available to the worker nodes for the next iteration.

%% file: Evaluation.tex
\section{Performance Evaluation of \our}\label{evaluation}

As discussed in Section~\ref{introduction}, the existing state-of-the-art works address fault tolerance by ensuring low failure probability in the edge ecosystem. However, no work tries to address the issue when a set of worker nodes fail. In this section, we compare \our with a scenario that ignores the failure of nodes and computes the global model using the remaining selected nodes~\cite{bonawitz2019towards} running vanilla FedAvg.

\subsection{Experiment Setup}
We evaluate \our under two experimental setups. First is in a simulation environment with PySyft as we have used in the previous experiments. Second is in a prototype environment with eight Raspberry Pi4~\cite{rpi} devices acting as worker nodes and an Ubuntu server acting as the aggregator. The details follow.

\subsubsection{Simulation Setup} We setup a simulation environment with Pysyft~\cite{pysyft}. The simulations are run on an Ubuntu 20.04 system~\cite{ubuntu} that has 12 GB RAM, an octa-core 1.5 GHz processor, and 16 GB Nvidia T4 GPU. For this set of experiments, we use the value of $\mathcal{K}$ as 100 and run the FL model on the MNIST dataset with MCNN. The data and bandwidth is synthetically assigned to each node following a Normal distribution~\cite{feller1968introduction}. This ensures that a fraction of the worker nodes have high data volume and a similar fraction have low data volume. We run the experiments for $20\%$, $40\%$, and $60\%$ of the worker nodes failing. It should be noted that the failing worker nodes are the ones that are selected as per the $\mathcal{S}-based$ selection strategy.

\subsubsection{Prototype Setup} We design the prototype environment with eight Raspberry Pi4 devices having 4 GB RAM and a quad-core 1.5 GHz processor. Two RPis have a storage size of 8 GB, two RPis have a storage of 4 GB and the remaining four RPis have a storage of 2 GB. The aggregator is run on a Ubuntu 20.04 system, with an 8 GB RAM and octa-core 1.5 GHz processor. Four RPis (8 GB, 4 GB and two 2 GB) are connected to the aggregator over a WiFi network having a bandwidth of 10 Mbps and the other four are connected through an Ethernet line of 100 Mbps.

\subsection{Simulation Results on \our}

We plot the $\mathcal{C}$ it takes for \our and FedAvg to reach a stable $\mathcal{A}$ when $20\%$, $40\%$, and $60\%$ of the selected nodes fail.

\begin{figure}[!ht]
  \centering
  \includegraphics[width=1\linewidth]{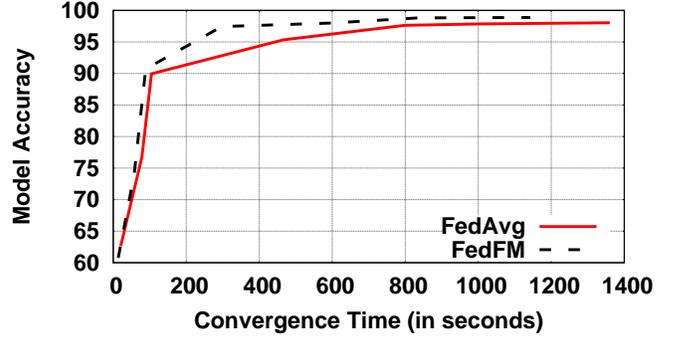}
  \caption{Convergence Time of FedAvg and \our with $20\%$ failed worker nodes.}
  \label{fig:mit_sim_20}
\end{figure}

\begin{figure}[!ht]
  \centering
  \includegraphics[width=1\linewidth]{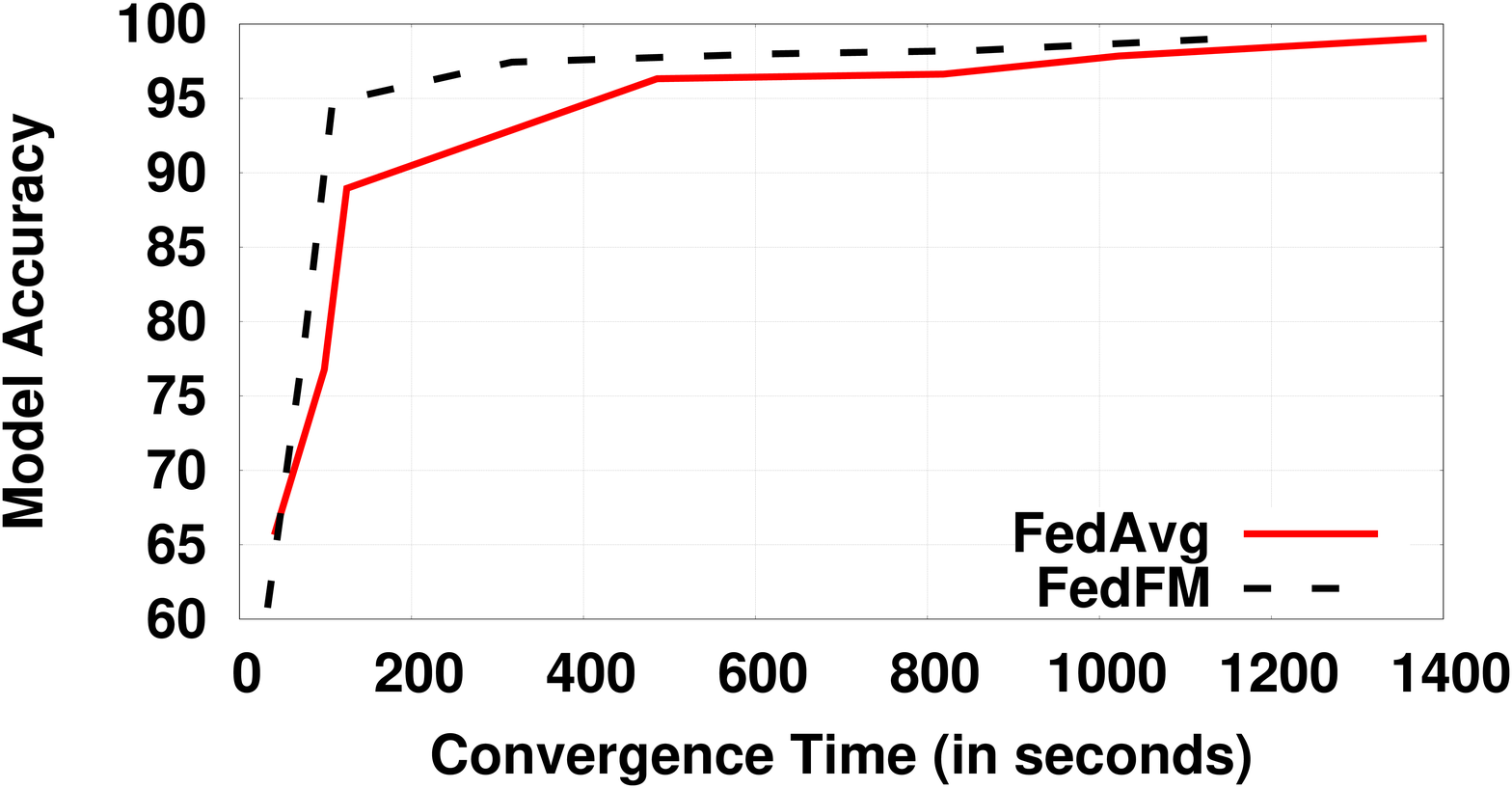}
  \caption{Convergence Time of FedAvg and \our with $40\%$ failed worker nodes.}
  \label{fig:mit_sim_40}
\end{figure}

\begin{figure}[!ht]
  \centering
  \includegraphics[width=1\linewidth]{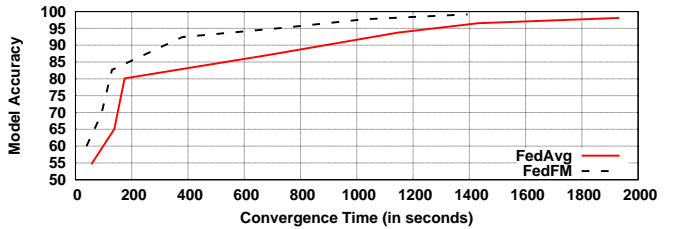}
  \caption{Convergence Time of FedAvg and \our with $60\%$ failed worker nodes.}
  \label{fig:mit_sim_60}
\end{figure}

In all three scenarios, i.e.,  Figure~\ref{fig:mit_sim_20},~\ref{fig:mit_sim_40},~\ref{fig:mit_sim_60} we observe that \our achieves higher $\mathcal{A}$ than FedAvg. This is intuitive since having higher number of worker nodes improves $\mathcal{A}$. Moreover, a higher accuracy value is achieved earlier in \our than in FedAvg. \our achieves higher accuracy earlier owing to the fact that, unlike FedAvg, \our has a higher number of nodes even after failure.

\subsection{Results on the \our Prototype Environment}
Since $70\%$ nodes are being selected in an eight node environment, five worker nodes are available to contribute to the global model. For the prototype environment, we run an experiment with $50\%$ of the nodes failing (two failed worker nodes). In order to simulate a failure, we randomly disconnect two of the five nodes selected by the $\mathcal{S}-based$ selection strategy.

\begin{figure}[!ht]
  \centering
  \includegraphics[width=1\linewidth]{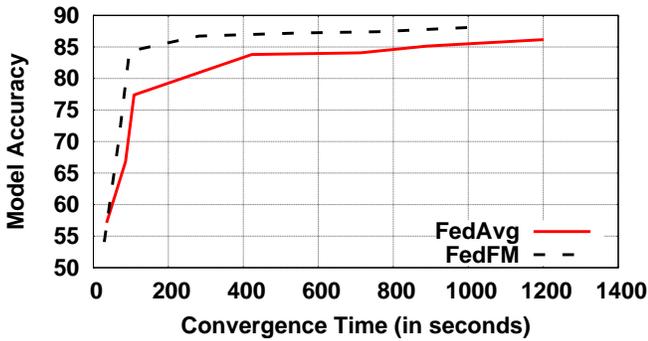}
  \caption{Convergence Time for FedAvg and \our when $50\%$ of nodes fail in the prototype environment.}
  \label{fig:mit_prot}
\end{figure}

Figure~\ref{fig:mit_prot} shows the value of $\mathcal{C}$ it takes for \our and FedAvg to reach a constant $\mathcal{A}$ when $50\%$ of the worker nodes fail. With only eight nodes, $\mathcal{A}$ is comparatively low for both models. However, we observe a similar trend as that of the simulation experiments for $\mathcal{A}$ and $\mathcal{C}$ in the prototype environment experiments.

These results validate our claim that fault mitigation is crucial for any Federated Learning ecosystem. With \our we are able to improve the \textit{Convergence Time} and \textit{Model Accuracy} for an FL technique.

%% file: Discussion.tex
\section{Discussion}\label{discussion}
In this paper, we provide a detailed experimental analysis of the impact of the number of worker nodes with Federated Learning technique. We observe how worker node failure affects the technique. We propose a new Federated Learning technique that builds upon FedAvg to provide a fault mitigation strategy. There are a few key aspects that need further analysis.

\subsection{Decentralized Federated Learning}
There is no fixed aggregator in a decentralized FL framework; instead, the worker nodes coordinate to generate the global model. This strategy removes the single-point of failure aggregator server. The network topology is crucial for the overall performance of the edge ecosystem.

In such a framework, $\mathcal{S}-based$ selection strategy is unlikely to work since several other factors come into play. For example, the computation power is responsible for training the model and aggregation using local models from other worker nodes. Similarly, the available bandwidth impacts the retrieval of local models from other worker nodes. Furthermore, another contributing parameter is the number of connected worker nodes in the network topology.

In the face of failure, the FL aggregation would need an upgrade. The enhanced FL algorithm should consider how a worker node should act when other worker nodes fail. The enhanced FL algorithm also needs to consider that when a particular node fails, it could render other nodes unreachable depending on the network topology.

\subsection{Dynamic Network Architecture}
This paper considers a network architecture where all the worker nodes and the aggregator are static, i.e., the network topology never changes. There could be network architectures where this might not be the case. Nodes could join and leave edge ecosystem dynamically. Furthermore, the nodes could have a dynamic system and network parameters. For example, based on the mobility of a node the network bandwidth could be a variable parameter. Similarly, considering smartphones, the storage available can also vary dynamically. Such a dynamic behaviour would affect both the selection and the failure mitigation strategies. 

\subsection{Incorporating Fairness in Node Selection}
The selection strategy that we use awards high score nodes, but unless there is a failure scenario, neglects low score nodes in the network. Fairness based strategies ensure that all nodes could be involved in the aggregation process. Fairness also guarantees that the global model does not miss the data exclusively available at the low score nodes. The selection strategy should be privacy-preserving and not decide fairness based on parameters that could affect data privacy.

\subsection{Homomorphic Encryption}
Homomorphic Encryption (HE)~\cite{armknecht2015guide} is a technique that enables computation on encrypted data without the need to decrypt it. Utilizing HE in an FL technique would have three advantages. First, the local models could be sent to the aggregator in a privacy-preserving manner; hence, the aggregator does not need to learn the exact weights. Second, the node selection strategy could be improved to incorporate encrypted data characteristics from the local nodes that ensures privacy. Finally, replication strategies could be employed for failure mitigation. The worker nodes could share encrypted copies of their data with other nodes, which could be used if the node fails anytime in the future.

%% file: Conclusion.tex
\section{Conclusion}~\label{conclusion}
AI on edge demands that we run complex data training tasks on edge itself. Addressing these demands along with privacy concerns is where Federated Learning comes in. With several heterogeneous edge devices contributing to the learning process, understanding the impact of these devices on the training and how the training process is hampered when these devices fail is crucial for an efficient FL technique. In this paper, we perform experimental analysis to understand essential aspects linked to the number of edge devices in the edge ecosystem, impact of the worker node count on two crucial FL parameters, the \textit{Model Accuracy} and \textit{Convergence Time}, and how these parameters are affected when the worker nodes fail. Following this, we provide a new Federated Learning technique that selects the optimal number of worker nodes for an efficient FL model. The FL technique employs a mitigation strategy when any of the worker nodes fail. The key takeaways from the paper are: (i) the number of worker nodes plays an integral part in the efficiency of an FL technique and is dependent on the learning model's architecture, (ii) not all nodes in the network are required for an efficient FL model, empirically, $70\%$ of the total nodes would perform as well as all the available nodes, (iii) having a specific number of working nodes in the network is not the same as having the same number of nodes post-failure as the failed nodes could have exclusive data samples, thus hindering the model performance, (iv) \our improves upon the existing FL techniques by employing fault mitigation strategies.

There exist key aspects which need further analysis. Decentralized and dynamic FL architectures would require that the selection and mitigation strategy be modified to address new challenges such as mobile worker nodes, unreachable worker nodes due to node failures, transmitting and receiving network delays, and varying system and network characteristics. Incorporating fairness in node selection is another area that needs to be looked into as it ensures that all worker nodes are able to participate in aggregation, and hence the learning model does not miss out on the data available at the weaker nodes. Homomorphic Encryption is another strategy that could help strengthen privacy and ensure fairness when employed with FL. HE could also be used for data replication in the network to open new directions for fault mitigation techniques.